\documentclass[letterpaper]{article}

\usepackage{aaai20}
\usepackage{times, amsmath}
\usepackage{helvet}
\usepackage{courier}
\usepackage[hyphens]{url} 
\usepackage{graphicx}
\newcommand{\bds}[1]{\boldsymbol{#1}}

\usepackage{graphicx}
\title{Deep Clustering with Measure Propagation}
\author{
Minhua Chen,\textsuperscript{\rm 1}
Badrinath Jayakumar,\textsuperscript{\rm 1} 
Padmasundari Gopalakrishnan,\textsuperscript{\rm 1}\\
\bf \Large Qiming Huang,\textsuperscript{\rm 1}
Michael Johnston,\textsuperscript{\rm 1}
Patrick Haffner,\textsuperscript{\rm 1}\\
\textsuperscript{\rm 1}Interactions LLC \\
\{mchen, bjayakumar, padma,
qhuang, \\
mjohnston, phaffner\}@interactions.com
}
\begin{document}

\maketitle
\setcounter{tocdepth}{1}

\section{Abstract}
Deep models have improved state-of-the-art for both supervised and unsupervised learning. For example, deep embedded clustering (DEC) has greatly improved the unsupervised clustering performance, by using stacked autoencoders for representation learning. However, one weakness of deep modeling is that the local neighborhood structure in the original space is not necessarily preserved in the latent space. To preserve local geometry, various methods have been proposed in the supervised and semi-supervised learning literature (e.g., spectral clustering and label propagation) using graph Laplacian regularization. In this paper, we combine the strength of deep representation learning with measure propagation (MP), a KL-divergence based graph regularization method originally used in the semi-supervised scenario. The main assumption of MP is that if two data points are close in the original space, they are likely to belong to the same class, measured by KL-divergence of class membership distribution. By taking the same assumption in the unsupervised learning scenario, we propose our Deep Embedded Clustering Aided by Measure Propagation (DECAMP) model. We evaluate DECAMP on short text clustering tasks. On three public datasets, DECAMP performs competitively with other state-of-the-art baselines, including baselines using additional data to generate word embeddings used in the clustering process. As an example, on the Stackoverflow dataset, DECAMP achieved a clustering accuracy of 79\%, which is about 5\% higher than all existing baselines. These empirical results suggest that DECAMP is a very effective method for unsupervised learning.

\section{Introduction}
Our society is generating a lot of short texts everyday. In social media such as Twitter and Facebook, short texts are posted on various aspects of everyday life. In the customer service domain, dialogues between customers and agents are transcribed and analyzed to find pattens in their interactions. More recently, virtual assistants such as Google Home have been responding to short inquiries from family users. The volume of the unlabeled texts is growing so huge that no human has the bandwidth to read through them, let alone label them. Hence it becomes necessary to use automated algorithms to analyze and organize short texts, documents and dialogues. As labeled data is often expensive to obtain, unsupervised learning methods which can cluster short texts in meaningful ways become very important.

In last decade, deep models have improved state-of-the-art for both supervised and unsupervised learning~\cite{imagenet}~\cite{gan}. For example, deep embedded clustering (DEC)~\cite{dec} has greatly improved unsupervised clustering performance over traditional methods such as K-means. By using stacked autoencoders~\cite{autoencoder} for representation learning, DEC can jointly learn the latent representation and the clustering, which is very attractive for analyzing high dimensional data such as image and text. Many deep clustering algorithms have been proposed since the initial demonstration of DEC. For example in IDEC~\cite{idec}, the autoencoder reconstruction loss is also included in the clustering process, while in the original DEC the decoder is discarded during clustering. In VaDE~\cite{vade}, the autoencoder is replaced by a variational autoencoder (VAE)~\cite{vae}, and a GMM prior is imposed in the latent space for joint representation learning and clustering. 

Consequently, it would be a natural idea to apply DEC to short text clustering. A very recent paper~\cite{acl} reported  encouraging results in this direction. However, one weakness of deep modeling is that the local neighborhood structure in the original space is not necessarily preserved in the latent space. Two data points in the original space may be mapped far apart through the deep encoders in DEC. Precisely due to this reason, the Laplacian autoencoder~\cite{lae} was proposed in the literature to remedy this issue. As local structure preserving is important for representation learning and clustering, we will focus on improving it in this paper. 

To preserve local geometry, various methods have been proposed in the supervised and semi-supervised learning literature (e.g., spectral clustering~\cite{sc} and label propagation~\cite{lp}) using graph Laplacian regularization~\cite{belkin}~\cite{niyogi}. A sparse graph is first constructed from the input space, so that nearby data points are connected in the graph. Then this graph is applied to the latent space through regularization to preserve the neighborhood structure. Besides graph regularization, another approach~\cite{gcn} which appeared recently uses the graph itself to generate latent representations through convolution and nonlinear operations. This graph convolutional network has been applied successfully to semi-supervised text classification problems. In the literature, there are some work on combining deep clustering with graph regularization, for example~\cite{lapdmm}~\cite{huawei}, which we will review in the experiment section.

In this paper, we combine the strength of deep representation learning with measure propagation (MP)~\cite{mp}, a KL divergence based graph regularization method originally used in a semi-supervised setting. The main assumption of MP is that if two data points are close in the original space, they are likely to belong to the same class, measured by KL divergence of class membership distribution. By taking the same assumption in the unsupervised learning scenario, we propose our Deep Embedded Clustering Aided by Measure Propagation (DECAMP) model. We evaluate DECAMP on short text clustering tasks. On three public datasets, DECAMP performs competitively with other state-of-the-art baselines, including baselines using additional data to generate word embeddings used in the clustering process. As an example, on the StackOverflow dataset, DECAMP achieves a clustering accuracy of 79\%, which is about 5\% higher than all existing baselines. These empirical results suggest that DECAMP is a very effective method for unsupervised learning.

Contributions in this work are summarized as follows:
\begin{enumerate}
     \item We combine the strength of deep clustering with graph regularization to preserve local structure of the data.
     \item We extend the application of measure propagation from semi-supervised to unsupervised learning.
     \item A special case of our algorithms provides a theoretical justification for the choice of the target distribution in DEC, which previously was considered as an empirical choice.
     \item Optimization of DEC with batch gradient algorithms could not be done end-to-end due to a frequency balancing term spanning across the entire data. We redefine this balancing constraint as a regularization term, and propose and new solution that can be fully optimized over a single mini-batch. This makes the DEC training more efficient.
     \item We improve the clustering performance of DEC without using additional data or external information, achieving state-of-the-art performance on short text clustering tasks.
\end{enumerate}
 

\section{Deep Embedded Clustering}\label{dec0}
Deep learning has improved both supervised and unsupervised learning greatly in the past decade. Recently there is a lot of work to apply deep models to clustering problems~\cite{survey}. Instead of clustering directly in the input space, deep clustering encodes the data and clusters in the latent space simultaneously. Thanks to the superb representation learning ability, deep clustering~\cite{infogan}~\cite{vade}~\cite{friendly} can yield substantial improvement over standard clustering methods such as K-means, Gaussian Mixture Models (GMM) and spectral clustering. 

One such deep clustering method is Deep Embedded Clustering (DEC)~\cite{dec} which we will focus on for this paper. In DEC, an autoencoder is pretrained on the input data $\{\bds{x}_i\}_{i=1}^{n}$ with a Mean-Squared Error (MSE) loss to reconstruct the input. This pretraining step is important, as early work on deep learning~\cite{science}~\cite{contrastive} has shown that it provides informative embedding and good initialization for further training. Then a clustering layer maps the latent representation $\{\bds{z}_i\}_{i=1}^{n}$ to the cluster predictive distribution $\{\bds{q}_i(\bds{\theta})\}_{i=1}^{n}$ via the following Student-t likelihood:
\begin{equation}\label{dec1}
     q_{ik}(\bds{\theta}) = \frac{(1+\|\bds{z}_i -\bds{\mu}_k\|^2)^{-1}}{\sum_{k'=1}^{K}(1+\|\bds{z}_i -\bds{\mu}_{k'}\|^2)^{-1}}
\end{equation}
Here $K$ is the total number of clusters, and the deep network parameter $\bds{\theta}$ includes both the encoder and the cluster centers $\{\bds{\mu}_k\}_{k=1}^{K}$. The encoder is initialized from the autoencoder pretraining process mentioned above, and the cluster centers are initialized via K-means clustering in the latent space. If the ground truth labels $\{\bds{p}_i\}_{i=1}^{n}$ were available, we could fine-tune the deep network parameter $\bds{\theta}$ using the KL divergence training loss (equivalent to cross-entropy loss) as follows:
\begin{equation}\label{dec2}
      \min_{\bds{\theta}} \ \frac{1}{n}\sum_{i=1}^{n}D_{KL}(\bds{p}_i || \bds{q}_i(\bds{\theta}) )
\end{equation}
However, for clustering problems, no such ground truth labels are available. In DEC, a self-training approach is taken to generate pseudo labels via the following equation:
\begin{equation}\label{dec3}
     p_{ik} = \frac{q^2_{ik}(\bds{\theta})/(\frac{1}{n}\sum_{j=1}^{n}q_{jk}(\bds{\theta}))}{\sum_{k'=1}^{K}q^2_{ik'}(\bds{\theta})/(\frac{1}{n}\sum_{j=1}^{n}q_{jk'}(\bds{\theta}))}
\end{equation}
This procedure will generate sharpened and balanced pseudo labels, due to the square operator and the normalization over $\frac{1}{n}\sum_{j=1}^{n}q_{jk}(\bds{\theta})$. The DEC workflow can be summarized as follows:
\begin{enumerate}
     \item Initialize the deep network via autoencoder pretraining and K-means clustering.
     \item Generate pseudo labels according to (\ref{dec3}).
     \item Train the deep network with fixed pseudo labels using loss in (\ref{dec2}).
     \item Repeat the above two steps until convergence. Output predictive distribution in (\ref{dec1}).
\end{enumerate}

\section{Deep clustering with Graph Regularization}
One weakness of DEC is that the local structure in the original space is not necessarily preserved in the latent space. To be more specific, if two points are close in the original space, we would hope that they are mapped to the same cluster in the latent space, which is a common assumption in manifold learning~\cite{lle}~\cite{isomap}. However, this is not necessarily the case for DEC, as it could map them far apart through the multi-layer deep encoder. To preserve the local neighborhood geometry, we first construct a graph affinity matrix $\bds{W}=\{w_{ij}\}_{i,j=1}^{n}$ in the original space, where $w_{ij}$ is nonnegative, and it is nonzero only if $\bds{x}_i$ and $\bds{x}_j$ are neighbors according to some affinity metric. This sparse matrix $\bds{W}$ provides pair-wise similarity information which could guide the deep clustering process. Hence we propose to optimize the following objective function for deep clustering:
\begin{equation}
\begin{split}
\label{mp1}
     \min_{\bds{p},\bds{\theta}} \ & C_{KL}(\bds{p},\bds{\theta}) = \frac{1}{n}\sum_{i=1}^{n}D_{KL}(\bds{p}_i || \bds{q}_i(\bds{\theta}) ) \\
    &  - \xi H(\frac{1}{n}\sum_{j=1}^{n}\bds{q}_{j}(\bds{\theta}))  + \frac{\lambda}{n} \sum_{i=1}^{n}H(\bds{p}_i)  \\&+ \frac{\nu}{n}\sum_{i=1}^{n}\sum_{j=1}^{n}w_{ij}D_{KL}(\bds{p}_i || \bds{p}_j)
\end{split}
\end{equation}
where each $\bds{p}_i$ is restricted to be a probability distribution over $K$ clusters, and $H(\cdot)$ is the Shannon entropy. This objective function is inspired by the work of~\cite{mp}, but a key difference is that here we are dealing with unsupervised instead of semi-supervised learning. We explain the functionality of each term in detail as follows:
\begin{enumerate}
     \item The first term is the same as the KL divergence loss (\ref{dec2}) in DEC. 
     \item The second term is a balancing regularization for clustering. We would like the average predictive distribution $\frac{1}{n}\sum_{j=1}^{n}\bds{q}_{j}(\bds{\theta})$ close to a uniform distribution, so that all clusters can be occupied to avoid degenerate solutions.
     \item The third term is the entropy regularization on the pseudo labels, to make the label distribution sharp and unequivocal, as low entropy implies high confidence. 
     \item The fourth term is the graph regularization to preserve local structure. If the affinity weight $w_{ij}$ is large (i.e., $\bds{x}_i$ and $\bds{x}_j$ are close to each other), the objective function will drive $D_{KL}(\bds{p}_i || \bds{p}_j)$ small, which means that they are likely to be mapped to the same cluster in the latent space. In this way, the graph affinity information in the original space will guide the deep clustering process.
\end{enumerate}

\section{Optimization via Measure Propagation}\label{omp}
We take an alternating minimization~\cite{csiszar} approach to solve (\ref{mp1}). First, given fixed deep network parameter $\bds{\theta}$ we solve the pseudo labels $\{\bds{p}_i\}_{i=1}^{n}$, and then given the generated pseudo labels we optimize the deep network.

\subsection{Pseudo Label Generation}
As no analytical solution exists for $\bds{p}_i$, we make use of the Measure Propagation (MP) method proposed in~\cite{mp}. We first relax (\ref{mp1}) to the following form:
\begin{equation}\label{mp2}
\begin{split}
     \min_{\bds{p}, \bds{r}, \bds{\theta}} \ & C_{MP}(\bds{p},\bds{r},\bds{\theta}) =  \frac{1}{n}\sum_{i=1}^{n}D_{KL}(\bds{p}_i || \bds{q}_i(\bds{\theta}) ) 
     \\& - \xi H(\frac{1}{n}\sum_{j=1}^{n}\bds{q}_{j}(\bds{\theta})) + \frac{\lambda}{n} \sum_{i=1}^{n}H(\bds{p}_i) + \\
      & \frac{\nu}{n}\sum_{i=1}^{n}\sum_{j=1}^{n}w_{ij}D_{KL}(\bds{p}_i || \bds{r}_j) + \frac{\nu}{n}\sum_{j=1}^{n}\alpha D_{KL}(\bds{p}_j || \bds{r}_j)
\end{split}\end{equation}
Here we introduced auxiliary labels $\{\bds{r}_i\}_{i=1}^{n}$ to make the objective function more tractable. It is easy to verify that 
\begin{equation}\label{mp3}
      \lim_{\alpha\rightarrow\infty}\min_{\bds{p}, \bds{r}, \bds{\theta}} \ C_{MP}(\bds{p},\bds{r},\bds{\theta}) 
           = \min_{\bds{p},\bds{\theta}} \ C_{KL}(\bds{p},\bds{\theta})
\end{equation}      
which bridges the optimization in (\ref{mp1}) and (\ref{mp2}). In practice we set a finite value for $\alpha$. More theoretical analysis on measure propagation could be found in~\cite{mp}, which could provide justifications for the above relaxation with a finite $\alpha$. By adding self-linking edges to the graph
\begin{equation}\label{mp4}
     \widetilde{\bds{W}} = \bds{W} + \alpha\bds{I}_n
\end{equation}
we could rewrite (\ref{mp3}) as 
\begin{equation}\label{mp5}
\begin{split}
     \min_{\bds{p}, \bds{r}, \bds{\theta}} \ & C_{MP}(\bds{p},\bds{r},\bds{\theta}) = \frac{1}{n}\sum_{i=1}^{n}D_{KL}(\bds{p}_i || \bds{q}_i(\bds{\theta}) )
      \\& - \xi H(\frac{1}{n}\sum_{j=1}^{n}\bds{q}_{j}(\bds{\theta})) + \frac{\lambda}{n} \sum_{i=1}^{n}H(\bds{p}_i)  \\& + 
      \frac{\nu}{n}\sum_{i=1}^{n}\sum_{j=1}^{n}\widetilde{w}_{ij}D_{KL}(\bds{p}_i || \bds{r}_j) 
\end{split}
\end{equation}
Now we can resort to alternating minimization again to solve $\bds{p}_i$ and $\bds{r}_j$ analytically as
\begin{eqnarray} 
     \label{mp6}
     p_{ik} \!\!\!\! &=& \!\!\!\! \frac{\exp( (\log q_{ik}(\bds{\theta}) + \nu\sum_{j=1}^{n}\widetilde{w}_{ij} \log r_{jk}) / \widetilde{\lambda}_i )}
     {\sum_{k'=1}^{K} \exp( (\log q_{ik'}(\bds{\theta}) + \nu\sum_{j=1}^{n}\widetilde{w}_{ij} \log r_{jk'}) / \widetilde{\lambda}_i )} \\
     \label{mp7}
     r_{jk} &=& \sum_{i=1}^{n}\widetilde{w}_{ij} p_{ik} / \sum_{i=1}^{n}\widetilde{w}_{ij}
\end{eqnarray} 
where $\widetilde{\lambda}_i = (1-\lambda) + \nu\sum_{j=1}^{n}\widetilde{w}_{ij}$. These two equations are iterated until convergence, which constitutes measure propagation for pseudo label generation. The first equation aggregates evidence from the deep network prediction and the auxiliary labels, and the second equation propagates the pseudo labels through the graph to update the auxiliary labels. Notice that each iteration is done simultaneously for all samples, with a computation complexity of $O(n\cdot m\cdot K)$ where $m$ is the number of nearest neighbors in the sparse graph. Since $m<<n$ and $K<<n$, each iteration of measure propagation is of linear complexity with respect to the sample size. More details on constructing the $\widetilde{\bds{W}}$ matrix can be found in the experiment section.

\subsection{Deep Network Optimization}
Given the pseudo labels $\{\bds{p}_i\}_{i=1}^{n}$ from measure propagation, we would like to optimize the deep network. By focusing only on terms related to $\bds{\theta}$ in (\ref{mp1}), we obtain the following loss function:
\begin{equation}\label{mp8}
     \min_{\bds{\theta}} \ \frac{1}{n}\sum_{i=1}^{n}D_{KL}(\bds{p}_i || \bds{q}_i(\bds{\theta}) ) - \xi H(\frac{1}{n}\sum_{j=1}^{n}\bds{q}_{j}(\bds{\theta}))
\end{equation}
which is equivalent to a standard cross-entropy loss plus a balancing regularization to avoid degenerate solutions~\cite{depict}. As the regularization requires averaging over the full dataset, which is unavailable during the mini-batch based training, we approximate it within each mini-batch separately, i.e., replacing $\frac{1}{n}\sum_{j=1}^{n}\bds{q}_{j}(\bds{\theta})$ with $\frac{1}{|B|}\sum_{j\in B}\bds{q}_{j}(\bds{\theta})$. Similar approximation is used in~\cite{imsat}.

We call our full algorithm Deep Embedded Clustering Aided by Measure Propagation (DECAMP), which is summarized as follows:
\begin{enumerate}
     \item Initialize the deep network via autoencoder pretraining and K-means clustering.
     \item Generate pseudo labels by measure propagation, iterating between (\ref{mp6}) and (\ref{mp7}).
     \item Train the deep network with fixed pseudo labels using loss in (\ref{mp8}).
     \item Repeat the above two steps until convergence. Output predictive distribution in (\ref{dec1}).
\end{enumerate}
The reader can compare it with DEC in Section \ref{dec0}, to see the differences in workflow.

\begin{figure}
  \centering
      \includegraphics[width=0.4\textwidth]{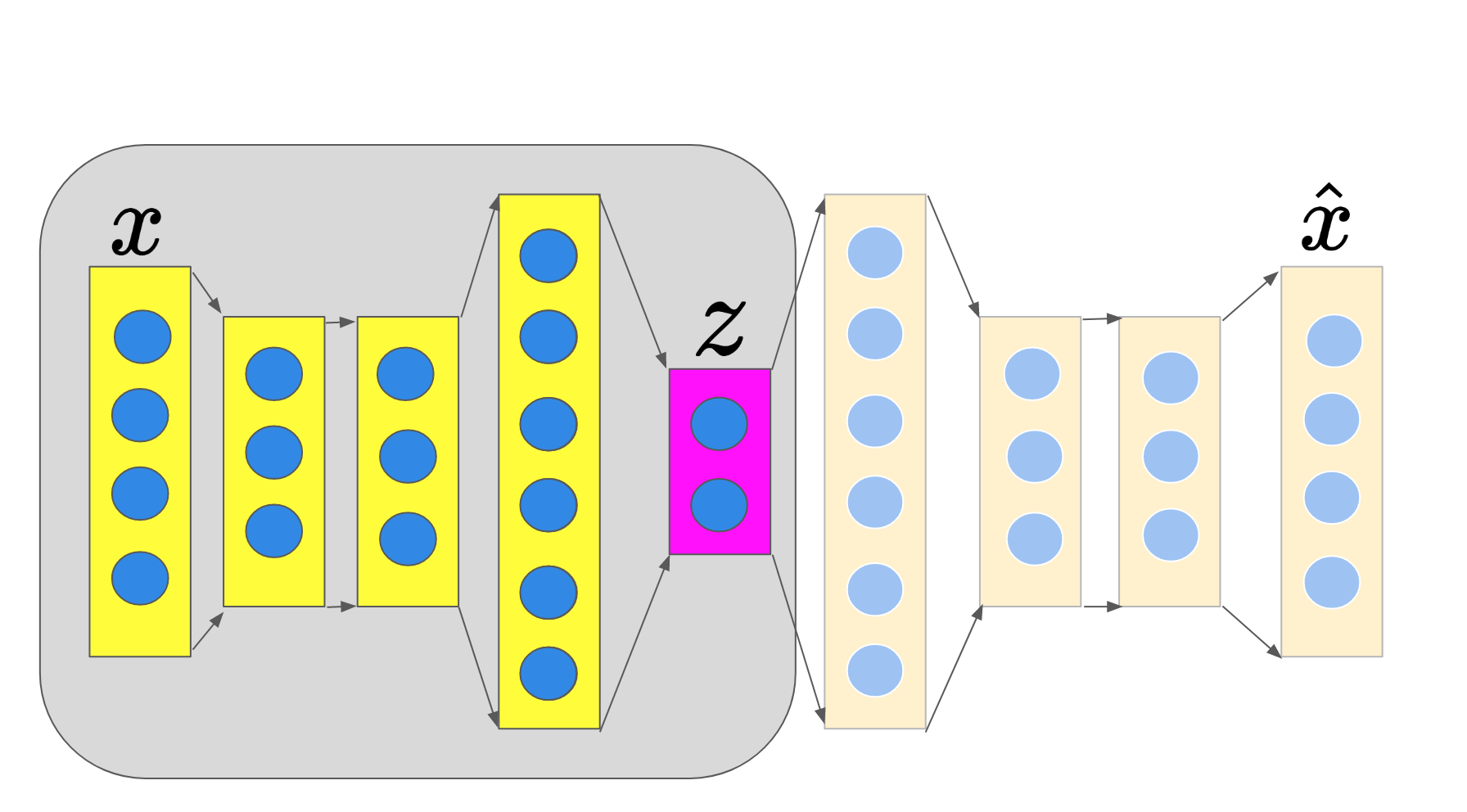}
      \includegraphics[width=0.4\textwidth]{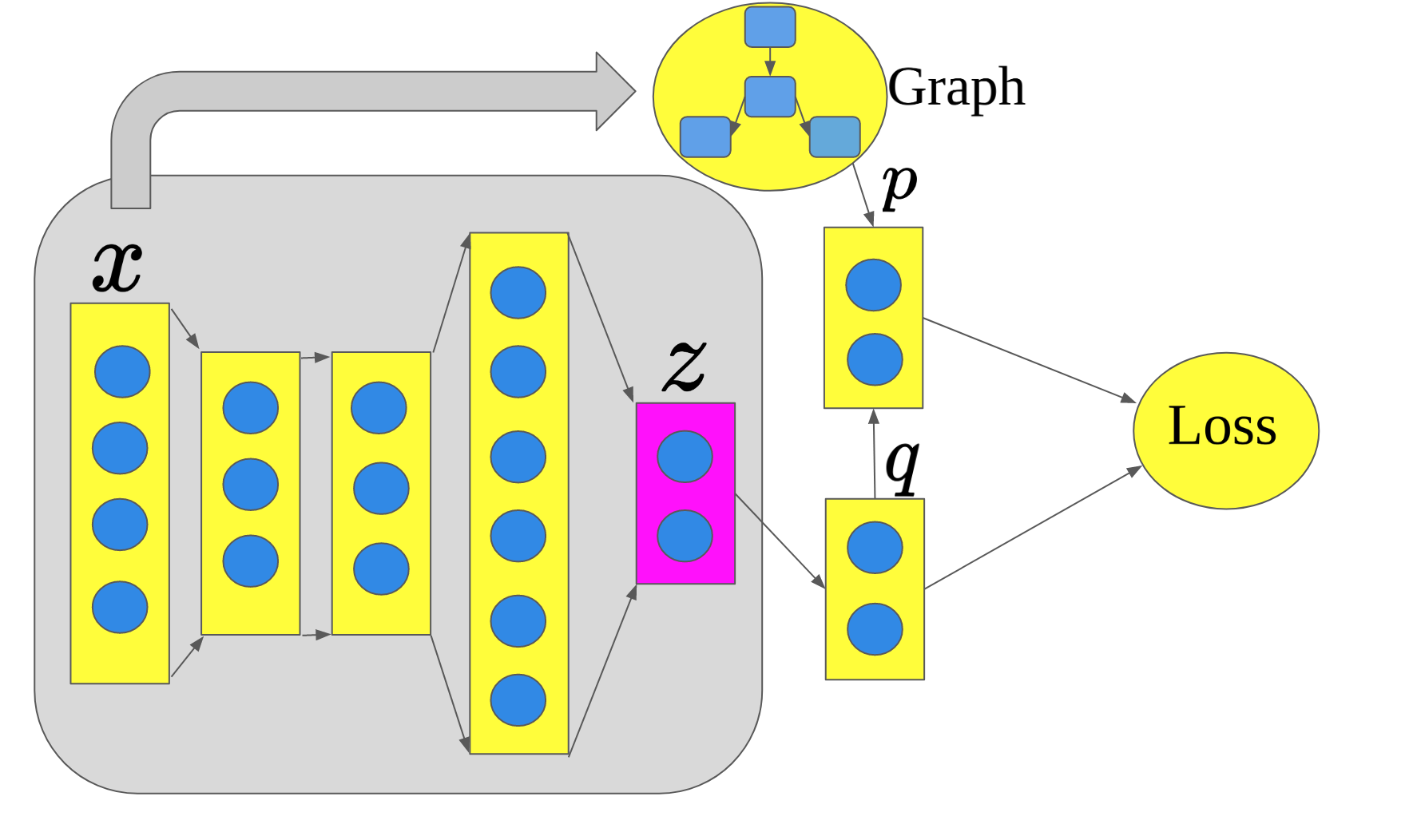}
  \caption{Illustration of the DECAMP algorithm. The first row is autoencoder pretraining, and the second row is the main clustering process using deep learning and measure propagation.}
\end{figure} 

\section{Special Case of $\nu=0$}
An an ablation study and sanity check, when $\nu=0$ in (\ref{mp1}), our algorithm should reduce to one similar to the original DEC, as no graph affinity information is used any more. We make this connection in this section.

When $\nu=0$ in (\ref{mp1}), the pseudo labels can be solved analytically as 
\begin{equation}\label{sp0}
      p_{ik} = \frac{q_{ik}^{\frac{1}{1-\lambda}}(\bds{\theta})}{\sum_{k'=1}^{K}q_{ik'}^{\frac{1}{1-\lambda}}(\bds{\theta})}
\end{equation}
Instead of relying on alternating minimization to solve (\ref{mp1}), as was done in Section \ref{omp}, we can directly replace the above analytical solution back to (\ref{mp1}) to eliminate the pesudo labels. The result is an end-to-end loss function for deep clustering:
\begin{equation}\begin{split}\label{sp1}
      & \min_{\bds{\theta}} \ C_{E2E}(\bds{\theta}) \\
      &= - \frac{1-\lambda}{n}\sum_{i=1}^{n}\log \sum_{k=1}^{K}q_{ik}^{\frac{1}{1-\lambda}}(\bds{\theta})
      - \xi H(\frac{1}{n}\sum_{j=1}^{n}\bds{q}_{j}(\bds{\theta})) \\&= 
      \frac{\lambda}{n}\sum_{i=1}^{n} H_{\frac{1}{1-\lambda}}(\bds{q}_{i}(\bds{\theta})) - \xi H(\frac{1}{n}\sum_{j=1}^{n}\bds{q}_{j}(\bds{\theta})) 
\end{split}\end{equation}
where $ H_{\frac{1}{1-\lambda}}(\cdot)$ is the R\'enyi entropy~\cite{principe} with parameter $\frac{1}{1-\lambda}$. This loss function is related to the maximum mutual information criterion proposed in~\cite{perona}~\cite{imsat}, except that R\'enyi entropy instead of Shannon entropy appeared in our loss function above. We call this algorithm DECE2E, and would expect that it has similar clustering performance to the original DEC algorithm. However, a key difference is that in DECE2E we have a clear objective function which can be optimized end-to-end,  while DEC relies on an empirical pseudo label generation equation in (\ref{dec3}) and the algorithm has to be trained in an alternative manner. DECE2E is a by-product we obtained along the way, and we will compare it to the main DECAMP algorithm in the result section.

\section{Experiments}
\subsection{Data Description}
We evaluate our DECAMP algorithm on short text clustering tasks. We consider three public datasets: Searchsnippets, Stackoverflow and Biomedical. Searchsnippets is a dataset containing web search snippets in eight different domains. Stackoverflow is a collection of question titles from the Stack Overflow question and answer website. Biomedical is a snapshot of PubMed data provided by BioAsQ. A summary of data statistics can be found in Table \ref{stats}. More description of the data can be found in~\cite{stc2}~\cite{acl}. For each dataset, we first remove all stop words, and extract tf-idf features on the 2000 most frequent words. After applying $\ell_2$ normalization, we obtain the feature input vectors $\{\bds{x}_i\}_{i=1}^{n}$. 

\begin{table}\label{stats}
    \begin{tabular}{lllll}
        \hline
        \textbf{Dataset}        & \textit{C}  & \textit{T}     & \textit{N}    & \textit{$|$V $|$} \\ \hline
        SearchSnippets & 8  & 12.3k & 17.9 & 31k \\
        StackOverflow  & 20 & 20k   & 8.3  & 23k \\
        Biomedical     & 20 & 20k   & 12.9 & 19k \\ \hline
    \end{tabular}
    \caption{Statistics for the short text clustering datasets as used by REF: number of classes (\textit{C}), number of short texts (\textit{N}), average number of tokens per text (\textit{T}) and and vocabulary size ($|$\textit{V}$|$))}

\end{table}

\subsection{Experimental Configurations}
We set the number of clusters $K$ to be the ground truth number found in Table \ref{stats}. The decision of the number of clusters from the data itself is an important research topic~\cite{number}, and we leave it for future study. We use feed-forward layers in the autoencoder, with dimensions $d_{\bds{x}}-500-500-2000-d_{\bds{z}}$ for the encoder, and with reverse order for the decoder.  From the data preprocessing decription, we have $d_{\bds{x}}=2000$. For Stackoverflow and Biomedical we set $d_{\bds{z}}=10$, and run 100 epochs for autoencoder pretraining; for Searchsnippets we set $d_{\bds{z}}=100$, and run 1000 epochs for autoencoder pretraining. For both pretraining and the main clustering process, we use stochastic gradient descent (SGD) with step size 0.1 and momentum 0.9. The above settings are kept the same for both DEC and DECAMP to make the comparison fair.

The hyper-parameters in DECAMP are set as follows for all experiments: $\xi=1.0, \lambda=0.5, \nu=0.5$. The graph $\widetilde{\bds{W}}$ for measure propagation is constructed as follows. First we define the affinity graph $\bds{W}$ as
\begin{equation}\label{exp1}
      w_{ij} = \left\{
      \begin{split}
      &1 \ \ \textrm{if} \ \bds{x}_j  \textrm{ is among the top m nearest neighbors of } \ \bds{x}_i \\ &0 \ \ \textrm{else} 
      \end{split}\right.
\end{equation}
and we set $m=50$ in our experiments. The metric we use for nearest neighbor search is cosine similarity (i.e., $\frac{\bds{x}_i^{\top}\bds{x}_j}{\|\bds{x}_i\|_2\cdot \|\bds{x}_j\|_2}$). Then we add the self-linking edges $\widetilde{\bds{W}}=\bds{W} + \bds{I}_n$ as in equation (\ref{mp4}). Finally we normalize it as $\widetilde{\bds{W}} \leftarrow \widetilde{\bds{D}}^{-\frac{1}{2}} \widetilde{\bds{W}} \widetilde{\bds{D}}^{-\frac{1}{2}}$ according to suggestions in~\cite{gcn}, where $\widetilde{\bds{D}}=\textrm{diag}(\widetilde{\bds{W}}\bds{1}_n)$ contains the row sums of the graph matrix.

\subsection{Evaluation Metrics}
We evaluate the performance of DECAMP using three metrics: accuracy (ACC), normalized mutual information (NMI) and adjusted rand index (ARI). The score values of these metrics all belong to the range of $[0,1]$, and the higher the scores are, the better the clustering quality. All three metrics require access to the ground truth labels. Notice that the ground truth labels are not used in the clustering process for all the algorithms. They are only used in this evaluation phase when clustering is finished. 

Suppose the ground truth labels are $\bds{Y} = \{y_i\}_{i=1}^{n}$ and the predicted labels are $\bds{C}=\{c_i)\}_{i=1}^{n}$. In DEC and DECAMP we have $c_i = \textrm{arg}\max_{k} q_{ik}(\bds{\theta})$. The accuracy (ACC) measure is defined as
\begin{equation}\label{exp2}
      \textrm{ACC} = \frac{1}{n}\sum_{i=1}^{n}\delta(y_i = \textrm{map}(c_i))
\end{equation}
where $\delta(\cdot)$ is an indicator function, and $\textrm{map}(\cdot)$ represents the best $K\times K$ permutation to match the clustering result to the ground truth labels, through the Hungarian algorithm~\cite{papadi}. The normalized mutual information (NMI) is computed by
\begin{equation}\label{exp3}
      \textrm{NMI} = \frac{I(\bds{Y};\bds{C})}{\sqrt{H(\bds{Y})H(\bds{C})}}
\end{equation}
where $I(\bds{Y};\bds{C})$ is the mutual information between $\bds{Y}$ and $\bds{C}$, and $H(\cdot)$ is the Shannon entropy. The adjusted rand index (ARI) is the corrected-for-chance version of rand index, which is a similarity measure between two data clusterings. Detailed computation equations could be found in the literature, and we omit it here to save some space. 

\subsection{Result and Analysis}
First we review three baseline algorithms: STC2~\cite{stc2}, DEC-SIF\cite{acl} and LapDMM~\cite{lapdmm}. STC2 is an early and important baseline for short text clustering. It uses additional data (generic Wikipedia data, or in-domain abstracts and post contents) to pretrain word embeddings, and use these word embeddings in the convolutional neural network (CNN) for clustering the data. DEC-SIF is a DEC-based clustering algorithm. Instead of using the original tf-idf features, DEC-SIF uses weighted pooling of the pretrained word embeddings as feature input to DEC. LapDMM is a recently proposed model with graph Laplacian regularization on the posterior cluster distribution of a Dirichlet mixture model. There are two realizations of LapDMM: LapDMM-T and LapDMM-W, with the former one relying on the tf-idf feature to construct the graph, and the latter one relying on the word embeddings.

As DECAMP relies only on the input data for graph construction and clustering, it would be unfair to directly compare it with baselines using additional information or word embeddings derived from in-domain data. Nevertheless, we list all baseline models in the results below, to make our comparison more complete. Results on the three public datasets are listed in the tables. The results for GMM, DEC, DECE2E and DECAMP are obtained over 10 independent experimental runs, and results for other baselines are obtained from the corresponding papers. As other papers did not report ARI scores, we only report ARI for the above four algorithms which we run.

\begin{table*}[t]
\centering
  \begin{tabular}{ | c | c | c | c | c |}
    \hline
    \textbf{Model} & \textbf{Comment} & \textbf{Searchsnippets} & \textbf{Stackoverflow} & \textbf{Biomedical} \\ \hline
    GMM & Baseline  & 31.0 & 50.0 & 33.0 \\ \hline
    LapDMM-T & Baseline & 76.1 +\!/\!- 6.0 & 72.8 +\!/\!-  2.0 & - \\ \hline
    LapDMM-W & Baseline (using add. info) & 79.3 +\!/\!- 3.0 & 71.0 +\!/\!- 5.0 & - \\ \hline
    STC2 & Baseline (using add. info) & 77.0 +\!/\!- 4.1 & 51.1 +\!/\!- 2.9 & 43.0 +\!/\!- 1.3 \\ \hline
    DEC-SIF & Baseline (using add. info) & 77.1 +\!/\!- 1.1 & 59.8 +\!/\!- 1.9 & \textbf{54.8} +\!/\!- 1.0 \\ \hline
    DEC & Baseline & 76.9 +\!/\!- 0.9 & 74.7 +\!/\!- 2.0 & 41.6 +\!/\!- 1.0 \\ \hline
    DECE2E & Proposed & 77.8 +\!/\!- 1.0 & 74.1 +\!/\!- 1.8 & 42.5 +\!/\!- 1.0 \\ \hline
    DECAMP & Proposed & \textbf{80.6} +\!/\!- 0.8 & \textbf{79.7} +\!/\!- 2.1 & 45.3 +\!/\!- 0.9 \\ \hline
    \hline
\end{tabular}
    \caption{ACC (mean +/- std)}
   \label{acc}
\end{table*}

\begin{table*}[t]
\centering
  \begin{tabular}{ | c | c | c | c | c |}
    \hline
    \textbf{Model} & \textbf{Comment} & \textbf{Searchsnippets} & \textbf{Stackoverflow} & \textbf{Biomedical} \\ \hline
    GMM & Baseline  & - & - & - \\ \hline
    LapDMM-T & Baseline & 63.4 +\!/\!- 4.0 & 64.1 +\!/\!-  1.0 & - \\ \hline
    LapDMM-W & Baseline (using add. info) & 65.3 +\!/\!- 1.0 & 64.5 +\!/\!- 2.0 & - \\ \hline
    STC2 & Baseline (using add. info) & 62.9 +\!/\!- 1.7 & 49.0 +\!/\!- 1.5 & 38.1 +\!/\!- 0.5 \\ \hline
    DEC-SIF & Baseline (using add. info) & 56.7 +\!/\!- 1.0 & 54.8 +\!/\!- 1.9 & \textbf{47.1} +\!/\!- 0.8 \\ \hline
    DEC & Baseline & 64.9 +\!/\!- 1.1 & 75.3 +\!/\!- 0.6 & 37.7 +\!/\!- 0.8 \\ \hline
    DECE2E & Proposed & 67.0 +\!/\!- 1.2 & 75.0 +\!/\!- 0.6 & 38.3 +\!/\!- 0.8 \\ \hline
    DECAMP & Proposed & \textbf{69.5} +\!/\!- 1.0 & \textbf{75.6} +\!/\!- 0.7 & 40.5 +\!/\!- 0.4 \\ \hline
    \hline
\end{tabular}
    \caption{NMI (mean +/- std)}
   \label{nmi}
\end{table*}

\begin{table*}[t]
\centering
  \begin{tabular}{ | c | c | c | c | c |}
    \hline
    \textbf{Model} & \textbf{Comment} & \textbf{Searchsnippets} & \textbf{Stackoverflow} & \textbf{Biomedical} \\ \hline
    GMM & Baseline  & - & - & - \\ \hline
    DEC & Baseline & 60.3 +\!/\!- 1.2 & 42.5 +\!/\!- 0.7 & 19.8 +\!/\!- 1.0 \\ \hline
    DECE2E & Proposed & 61.8 +\!/\!- 1.5 & 41.1 +\!/\!- 0.7 & 21.5 +\!/\!- 1.4 \\ \hline
    DECAMP &  \hspace{1.6cm}  Proposed \hspace{1.6cm} & \textbf{66.3} +\!/\!- 1.3 & \textbf{60.6} +\!/\!- 0.9 & \textbf{26.2} +\!/\!- 0.7 \\ \hline
    \hline
\end{tabular}
    \caption{ARI (mean +/- std)}
   \label{ari}
\end{table*}

We have the following observations from the result tables:
\begin{enumerate}
      \item DECAMP performs better than DEC and DECE2E in all three datasets, across all three evaluation metrics.
      For example, we observed about 4\% ACC improvement for DECAMP across all three datasets. We provide detailed metric trajectories along training epochs in Figure ? for one experiment. As can be seen, while DEC and DECE2E's performance saturate in an early stage, DECAMP continues to improve along iterations till a higher performance level.
      \item DEC and DECE2E perform comparably across the datasets. This result was expected in our previous discuss, as neither of them use the graph affinity information in the original space. However, the main advantage of DECE2E is its theoretical elegance and training efficiency, as we can train it end-to-end through a unified loss function.
      \item Comparing with other state-of-art methods published on these three datasets, DECAMP is also among the best performers. For example, on Stackoverflow, DECAMP's ACC is about 5\% higher than all existing baselines, including methods using additional information during the clustering process. Notice that on the Biomedical dataset, DECAMP does not compete well with DEC-SIF, and we hypothesize that one reason might be the high-quality word embeddings in DEC-SIF derived from additional in-domain data. For baselines with no access to additional information source (such as DMM-T), we do see a big gain of DECAMP across the board.      
\end{enumerate}

\begin{figure}
  \centering
      \includegraphics[width=0.5\textwidth]{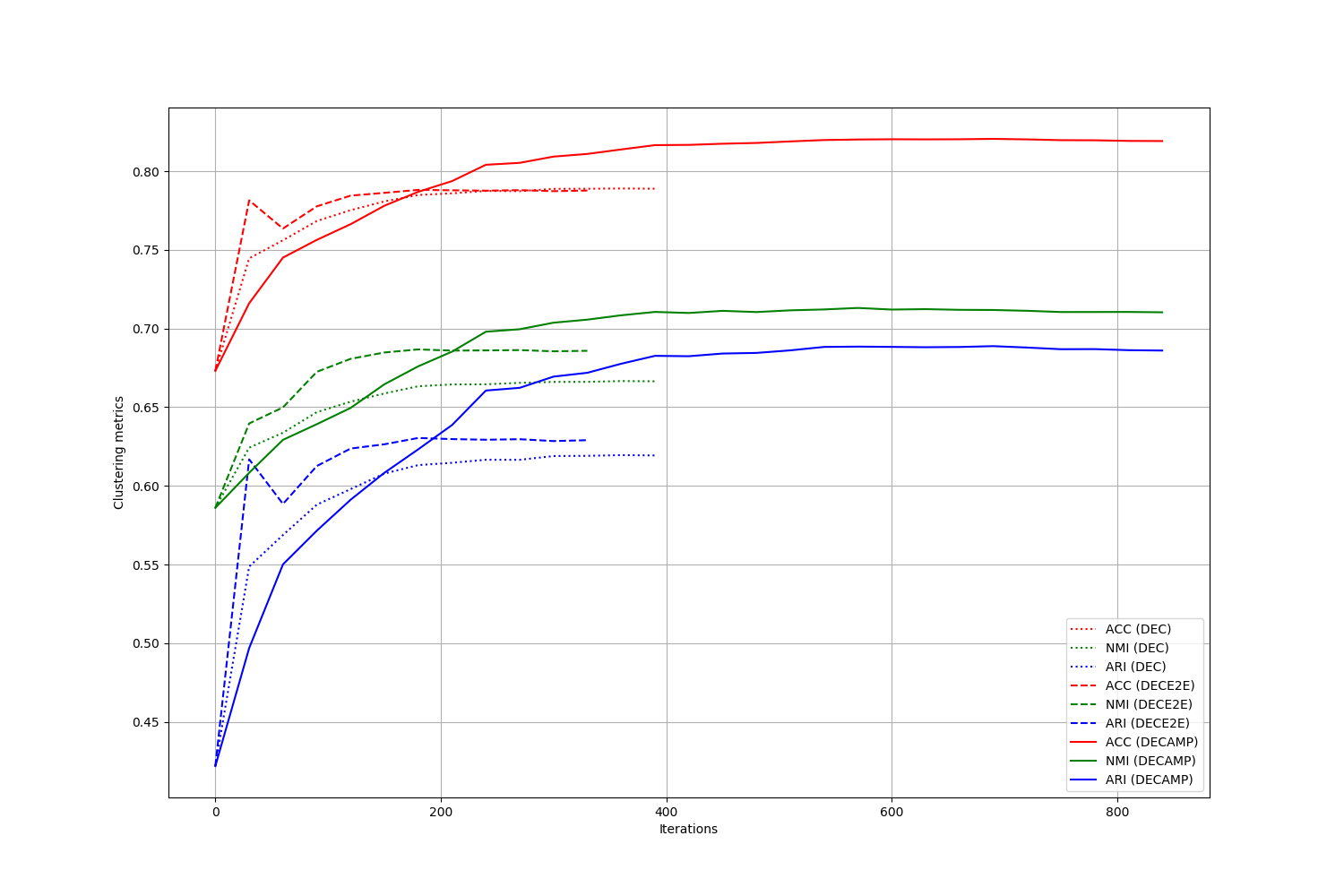}
  \caption{Searchsnippets}
\end{figure}   

\begin{figure}
  \centering
      \includegraphics[width=0.5\textwidth]{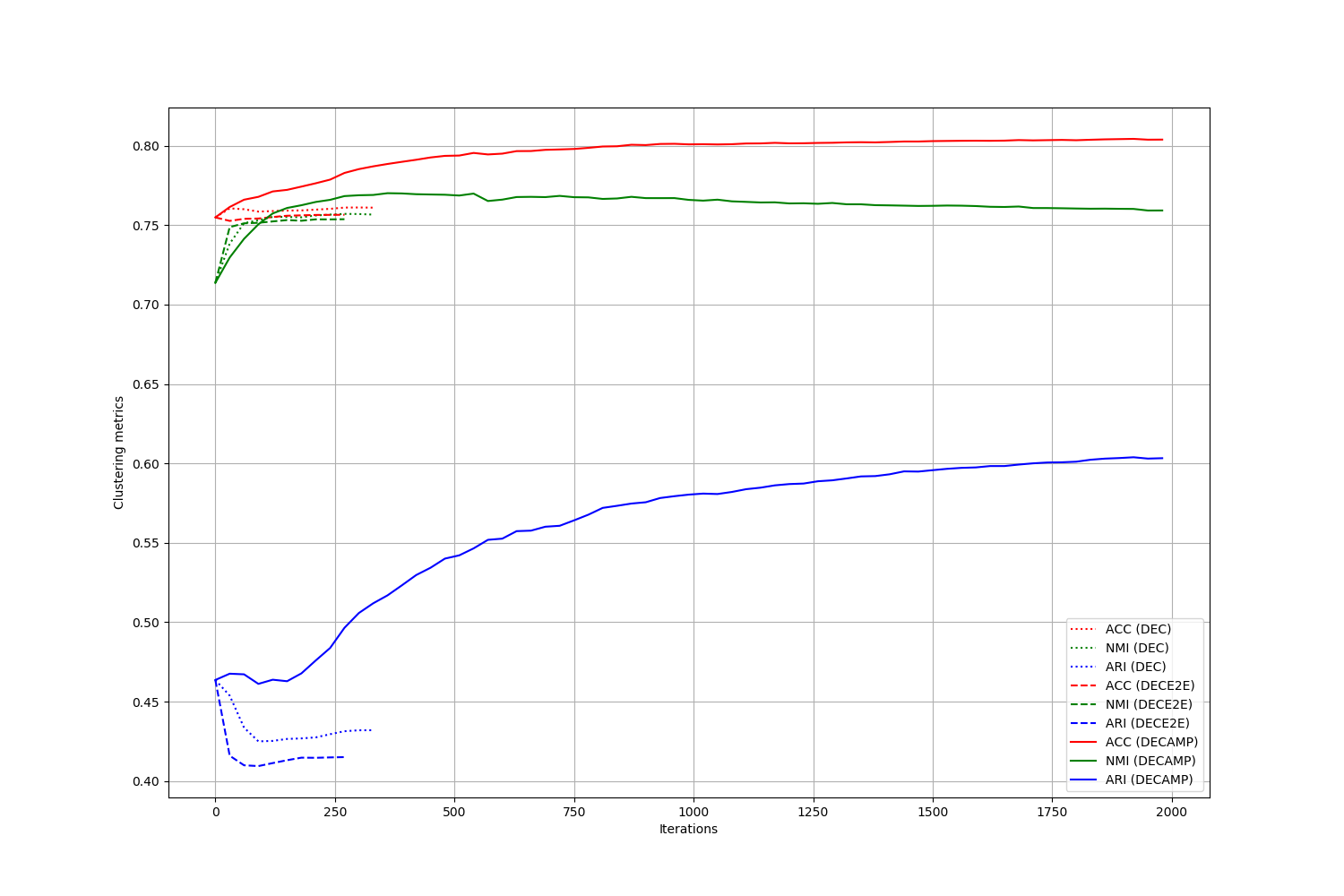}
  \caption{Stackoverflow}
\end{figure}   

\begin{figure}
  \centering
      \includegraphics[width=0.5\textwidth]{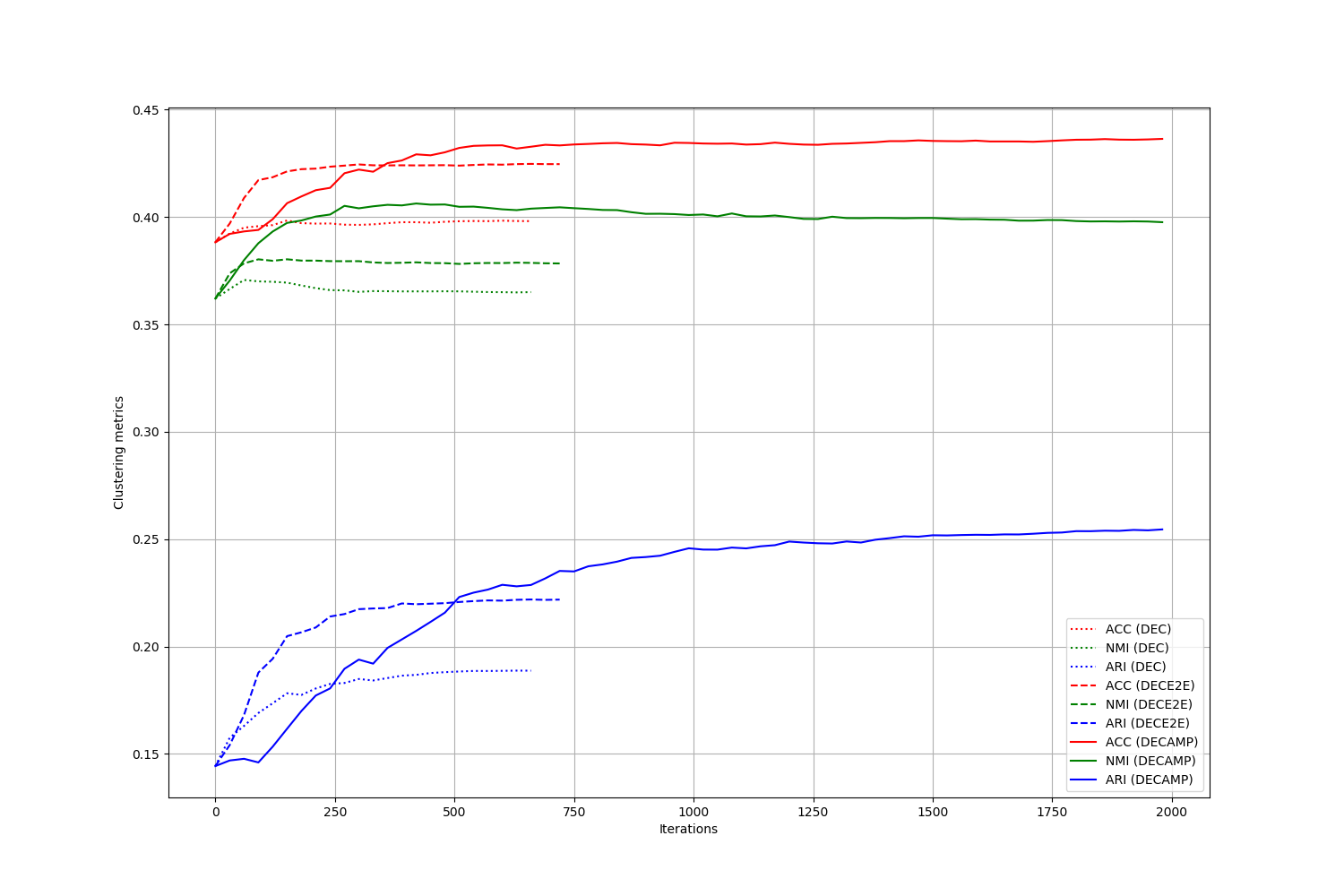}
  \caption{Biomedical}
\end{figure}

\section{Conclusion}
In this paper we proposed a new deep clustering method for short text analysis. By combining the strength of deep representation and measure propagation, our DECAMP algorithm uses the neighborhood affinity information to guide the clustering process, and achieves state-of-the-art performance on multiple public datasets. As a by-product, we obtained an end-to-end version of DEC which admits a unified objective function and allows more efficient training. As a future research direction, we will explore the possibility of encoding the measure propagation step as a neural network component, so that DECAMP can become a new kind of graph neural network~\cite{gnn} admitting end-to-end training.

\bibliographystyle{aaai}
\bibliography{DECAMP}

\end{document}